\documentclass[11pt]{article}
\ifdefined\XeTeXversion\else\pdfoutput=1\fi  % arXiv: pdflatex; tectonic (XeTeX) skips
\usepackage[utf8]{inputenc}
\usepackage[T1]{fontenc}
\usepackage{lmodern}
\usepackage{microtype}
\usepackage{amsmath,amssymb,amsthm}
\usepackage{booktabs}
\usepackage{graphicx}
\usepackage[round]{natbib}
\usepackage[margin=1.05in]{geometry}
\usepackage[colorlinks=true,linkcolor=blue,citecolor=blue,urlcolor=blue]{hyperref}
\usepackage{url}
\usepackage{tikz}
\usetikzlibrary{positioning,arrows.meta,shapes.geometric}
\tikzset{
  box/.style={draw=black!60, rounded corners=1.5pt, align=center, font=\footnotesize,
              inner sep=4pt, minimum height=8mm},
  rt/.style={box, fill=blue!8},
  ex/.style={box, fill=green!10},
  tn/.style={box, fill=orange!12},
  gt/.style={box, fill=black!6},
  arr/.style={-{Latex[length=2mm]}, draw=black!60},
  dec/.style={diamond, aspect=2.2, draw=black!60, fill=orange!12, align=center,
              font=\footnotesize, inner sep=1pt},
}
\newcommand{\code}[1]{\texttt{#1}}
\newtheorem{definition}{Definition}
\newtheorem{problem}{Problem}
\newcommand{\nPublic}{129}
\newcommand{\nSpecialist}{16}
\newcommand{\nBlendSpec}{32}
\newcommand{\nBlendBase}{22}
\newcommand{\nTournament}{27}
\newcommand{\nWithSpecialist}{48}

\newcommand{\nExperts}{43}
\newcommand{\nExpC}{19}
\newcommand{\nExpTirex}{13}
\newcommand{\nExpToto}{11}
\newcommand{\nBlendTwo}{23}
\newcommand{\nBlendThree}{30}
\newcommand{\nBlendFour}{1}
\newcommand{\nBlendAll}{54}
\newcommand{\nBlendMixed}{42}
\newcommand{\bestPoolName}{Toto 2.0 2.5B FT}
\newcommand{\bestPoolPos}{23}
\newcommand{\bestPoolRank}{33.8}
\newcommand{\worstPoolRank}{58.4}

\newcommand{\ourPos}{3}
\newcommand{\ourRank}{19.4}
\newcommand{\ourWqlRank}{24.6}
\newcommand{\ourWqlPos}{8}

\newcommand{\nEntries}{130}
\newcommand{\nBeatBest}{37}
\newcommand{\nWithinFive}{60}
\newcommand{\nWorseTen}{30}

\newcommand{\tourOnlyRank}{38.0}
\newcommand{\tourOnRankSub}{19.8}
\newcommand{\bestPoolRankSub}{28.1}

\newcommand{\nTournamentConfigs}{27}
\newcommand{\nSpecBeatsTour}{69}
\newcommand{\nOther}{70}
\newcommand{\medGainSpec}{2.0}

\title{TW3Cast: A Frozen Router of Lightly Fine-Tuned\\
Foundation Models for Time-Series Forecasting on GIFT-Eval,\\
Selected Entirely on the Training Split\thanks{Every leaderboard table, figure and number of this paper regenerates by script from the
released routing table, the released expert index and a dated snapshot of
the public per-configuration scores; see Section~\ref{sec:repro}.}}

\author{Nathan Thierry\\
TW3 Partners\\
\texttt{n.thierry@tw3partners.com}
\and
Andr\'e-Louis Rochet\\
TW3 Partners\\
\texttt{arochet@tw3partners.com}}

\date{September 2026}

\begin{document}
\maketitle

\begin{abstract}
TW3Cast is a time-series forecasting system that reaches position
\ourPos{} of \nEntries{} entries on the GIFT-Eval benchmark by mean MASE
rank, as of 2026-09-14. The two entries above it belong to the leaderboard's agentic category,
multi-step systems that use agents or language models to reason about,
generate or select forecasts. TW3Cast runs no agent and no language
model. Its selection is a table computed once on the training split and
then frozen, and its experts are public foundation models lightly
fine-tuned on those training splits. For each of the 97 dataset,
frequency and horizon configurations, the table serves one of four modes:
a specialist, which is a LoRA or full fine-tune of Chronos-2, TiRex or
Toto whose training data was cleaned and enriched by explicit rules; a
quantile blend that contains a specialist; a blend of base models; or a
selection tournament played on a backtest carved from the training
split. Every decision in the table was taken on that
backtest. A specialist is admitted the moment it beats the tournament
there, so a candidate costs a few megabytes and minutes of GPU time, and a
failed candidate changes nothing. Three guarded mechanisms protect the
selection from its own biases: a dual accuracy and calibration criterion,
an asymmetric margin against candidates that saw the series during
training, and conservative per-window gates. The selection rules
themselves were chosen inside a temporal meta-backtest. The best base
model served alone reaches a mean MASE rank of \bestPoolRank, the
tournament served on every configuration reaches \tourOnlyRank, and the
full router reaches \ourRank. The routing table, the expert index, the
pinned base-model revisions, the submitted score file and the dated
snapshot of the public scores are released, and every leaderboard number in this paper regenerates from them by one script.
\end{abstract}

\section{Introduction}
\label{sec:intro}

A practitioner who needs one forecast per data source now chooses among
several public foundation models, several sizes and several context
policies, with fine-tuning as a further option. On the GIFT-Eval benchmark
\citep{aksu2024gifteval}, which aggregates 97 configurations over 24
datasets, none of these models served alone reaches the top of the
leaderboard. The information that decides which model should speak on which data is in the recent history of
each source, and this paper builds a system around that observation.

TW3Cast is a router. For every configuration, identified by name, a frozen
table gives mixture weights over experts, and the forecast is the weighted,
sorted quantile average of the selected experts. The table is not learned
end to end. Each row is the outcome of an explicit decision procedure run
on a backtest carved out of the training split, and the row is then
frozen. At prediction time nothing is decided any more, apart from two
per-window guard rails with fixed thresholds. The router can therefore be
audited row by row, improved row by row, and shipped as two small files.

Four contributions organize the paper. First, a decision protocol in which
every choice is taken on a train-side backtest with a recency bias, and in
which the selection rules themselves are validated by a temporal
meta-backtest (Section~\ref{sec:protocol}). Second, an expert-construction
methodology in which light fine-tuning becomes competitive once the data
is prepared by explicit rules for per-series cleaning, dosed fusion of
sibling datasets, window sampling with robust residual normalization, and
a pinball loss on the nine evaluated quantiles
(Section~\ref{sec:experts}). Third, a set of
guarded selection mechanisms, a tournament with a dual accuracy and
calibration criterion and an asymmetric anti-memorization margin,
conservative per-window gates, and a correlation guard against misaligned
imports (Section~\ref{sec:tournament}). Fourth, an accretion property. Admission
requires beating the tournament on the backtest, so adding an expert never
degrades the system on its selection criterion, and a candidate costs
minutes. We report the wins and the configurations where every trained
candidate lost and was discarded (Sections~\ref{sec:results}
and~\ref{sec:deploy}). Inserted among the \nPublic{} public entries with complete
results on 2026-09-14, the system reaches position \ourPos{} of \nEntries{} by mean MASE rank
(Section~\ref{sec:results}).

\section{Related work}
\label{sec:related}

\paragraph{Time-series foundation models.} The expert pool builds on
public pretrained forecasters from three architecture families. Chronos-2
\citep{ansari2025chronos2} is a multivariate sequence model with joint
channel prediction, descended from Chronos \citep{ansari2024chronos}.
TiRex \citep{auer2025tirex} is an xLSTM-based univariate model, with a
second generation released as a model card \citep{tirex2card}. Toto
\citep{cohen2025toto} is a decoder model for observability data, of which
we adapt the 2.5B fine-tuned release \citep{toto2card} with LoRA
\citep{hu2021lora}. TimesFM \citep{das2024timesfm,timesfm25card} and a
public LoRA of Chronos-2 \citep{turkforecast} complete the tournament
pool. Other foundation forecasters, such as Moirai \citep{woo2024moirai},
Lag-Llama \citep{rasul2023lagllama} and the mixture-of-experts variants
Moirai-MoE \citep{liu2024moiraimoe} and Time-MoE \citep{shi2025timemoe},
are not in the pool but appear on the same leaderboard. 

\paragraph{Agentic entries.} The leaderboard's agentic category covers
multi-step systems that use agents or language models to reason about,
generate or select forecasts. The first entry above ours in Section~\ref{sec:results} combines two
published components. STRIDE \citep{ahamed2026stride} distills reasoning
traces into a lightweight language model whose hidden states are
projected into the encoder of a time-series foundation model, and Synapse
\citep{das2025synapse} arbitrates between several foundation models with
weights adjusted at inference. The second, EXAONE-Forecast-Agent, is
declared agentic by its authors, and we found no public description of
its pipeline. We do not characterize the other agentic entries. TW3Cast
shares with these systems the idea of routing among
public foundation models per data source, and differs in three choices: no language model anywhere, a routing
decided once on the training split and frozen, and experts obtained by
light fine-tuning.

\paragraph{Forecast combination.} That combining forecasts beats
committing to one goes back to \citet{bates1969combination}, and the
forecasting competitions have shown simple combinations to be strong
\citep{makridakis2020m4,makridakis2022m5,timmermann2006combinations}.
The forecast combination puzzle, in which estimated weights lose to equal
weights, has a simple explanation in the estimation noise of the weights
\citep{smith2009puzzle,claeskens2016puzzle}. Our router uses equal weights inside every blend and spends its
estimation budget on the choice of members.

\paragraph{Model selection by meta-learning.} Selecting or weighting
forecasters per series from features of the series is the FFORMS and
FFORMA line \citep{talagala2023fforms,montero2020fforma}, which itself
extends earlier meta-learning for forecast selection
\citep{lemke2010metalearning}. Those methods learn a selector across many
series. We select per configuration on a backtest of the configuration's
own history, with guard rails against the biases such a selection creates,
in particular the structural advantage of candidates that saw the series
during training. Mixture-of-experts routing \citep{shazeer2017moe} learns
the router jointly with the experts. We compute the routing table by
measurement and freeze it, which trades expressiveness for auditability
and zero inference-time overhead.

\paragraph{Backtesting for selection.} The choice between blocked temporal
splits and random splits for time-series model selection has a long
literature \citep{tashman2000outofsample,bergmeir2012crossvalidation,
cerqueira2020evaluating}. Our temporal meta-backtest applies the same choice one level up, to the
selection rule rather than to the model.

\section{Problem statement}
\label{sec:problem}

\begin{definition}[Configuration]
\label{def:config}
A configuration is a triple of dataset, sampling frequency and horizon
class in the benchmark. Each configuration provides a training split and a
sequence of official evaluation windows. GIFT-Eval has 97 configurations.
Each per-window forecast is scored by MASE \citep{hyndman2006mase} for
point accuracy and by the mean weighted quantile loss over the nine
deciles for calibration, a discrete approximation of the CRPS
\citep{gneiting2007scoring}. Leaderboard positions aggregate
per-configuration ranks.
\end{definition}

\begin{problem}[Per-configuration serving under zero test access]
\label{prob:main}
For each configuration, choose the forecaster to serve, a single model, a
quantile blend or a per-window policy over a pool, using only the training
split, so that the mean per-configuration rank over all 97 configurations
is minimized.
\end{problem}

\begin{definition}[Admission rule]
\label{def:admission}
Let $S(\cdot)$ denote the backtest score of a candidate under the dual
criterion of Section~\ref{sec:tournament}. A trained specialist is
designated for a configuration as soon as it beats the tournament's
champion on the backtest. Under comparable scores the specialist is
preferred. Otherwise the configuration keeps the tournament as its serving
mode.
\end{definition}

The admission rule makes improvement monotone by construction on the
selection criterion, and it prices a failed experiment at the cost of
training the candidate.

\section{The router}
\label{sec:router}

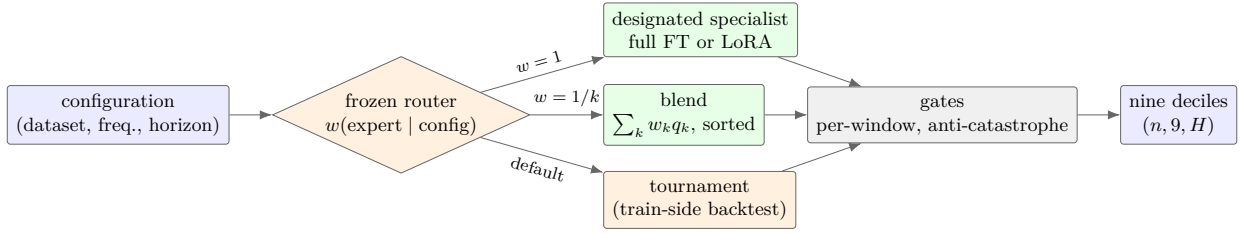
\begin{figure}[t]
\centering
\resizebox{\linewidth}{!}{%
\begin{tikzpicture}[node distance=5mm and 7mm]
\node[rt] (c) {configuration\\(dataset, freq., horizon)};
\node[dec, right=of c] (r) {frozen router\\$w(\text{expert}\mid\text{config})$};
\node[ex, right=12mm of r, yshift=14mm] (e) {designated specialist\\full FT or LoRA};
\node[ex, right=12mm of r] (b) {blend\\$\sum_k w_k q_k$, sorted};
\node[tn, right=12mm of r, yshift=-14mm] (t) {tournament\\(train-side backtest)};
\node[gt, right=of b] (g) {gates\\per-window, anti-catastrophe};
\node[rt, right=of g] (q) {nine deciles\\$(n, 9, H)$};
\draw[arr] (c) -- (r);
\draw[arr] (r) -- node[above, sloped, font=\scriptsize] {$w=1$} (e);
\draw[arr] (r) -- node[above, font=\scriptsize] {$w=1/k$} (b);
\draw[arr] (r) -- node[below, sloped, font=\scriptsize] {default} (t);
\draw[arr] (e) -- (g); \draw[arr] (b) -- (g); \draw[arr] (t) -- (g);
\draw[arr] (g) -- (q);
\end{tikzpicture}}
\caption{The frozen router. A configuration name resolves by table lookup
to a designated specialist, a quantile blend or the tournament mode.
Per-window gates then bound the tail risk, and the system outputs the nine
deciles for each evaluation window.}
\label{fig:router}
\end{figure}

The shipped system is a routing table \code{router.csv} with one row per
configuration and expert, giving a mixture weight, and an expert index
\code{experts.json} giving each expert's architecture family. Weights in a
row sum to one. Figure~\ref{fig:router} shows the four serving modes that
the table encodes. A designated specialist has weight one. A blend has
equal weights $1/k$ over its $k$ members, which may mix specialists and
base models. The reserved member \code{\_\_tournament\_\_} marks a
configuration whose forecast is produced by the train-side tournament of
Section~\ref{sec:tournament}. Inference is a table lookup, the experts'
forward passes and a sorted weighted average of quantiles.

\begin{figure}[t]
\centering
\includegraphics[width=0.95\linewidth]{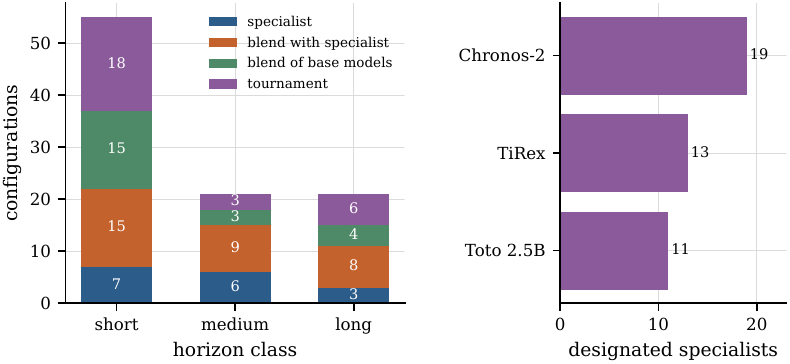}
\caption{Composition of the released router. Left, the serving mode of the
97 configurations by horizon class. Right, the architecture family of the
\nExperts{} designated specialists. Both panels are read from
\code{router.csv} and \code{experts.json} by \code{make\_figures.py}.}
\label{fig:composition}
\end{figure}

In the released table, \nSpecialist{} configurations are served by a single
specialist, \nBlendSpec{} by a blend that contains at least one specialist,
\nBlendBase{} by a blend of base models only, and \nTournament{} by the
tournament. The \nExperts{} designated specialists split into \nExpC{}
Chronos-2 experts, \nExpTirex{} TiRex experts and \nExpToto{} Toto experts.
Blends have two members in \nBlendTwo{} configurations, three in
\nBlendThree{} and four in \nBlendFour. Figure~\ref{fig:composition} gives
the breakdown by horizon class, and Appendix~\ref{app:router} lists the
full table.

\section{The decision protocol}
\label{sec:protocol}

\begin{figure}[t]
\centering
\resizebox{\linewidth}{!}{%
\begin{tikzpicture}[node distance=4mm and 9mm]
\node[rt] (tr) {training split};
\node[tn, right=of tr, yshift=9mm] (bt) {backtest\\100 to 600 windows per configuration,\\recency bias};
\node[rt, right=of tr, yshift=-9mm] (td) {remaining training data};
\node[ex, right=of td] (ft) {training\\full FT or LoRA};
\node[tn, right=of bt, xshift=6mm] (sel) {all decisions\\tournaments, admissions,\\blends, weights $w$};
\node[rt, right=of sel] (rt) {frozen router};
\draw[arr] (tr) -- (bt); \draw[arr] (tr) -- (td); \draw[arr] (td) -- (ft);
\draw[arr] (bt) -- (sel); \draw[arr] (ft) -- (sel); \draw[arr] (sel) -- (rt);
\end{tikzpicture}}
\caption{Every decision is taken on a backtest extracted from the training
split. The remaining training data trains the experts, and the resulting
router is frozen before any test window is read.}
\label{fig:protocol}
\end{figure}
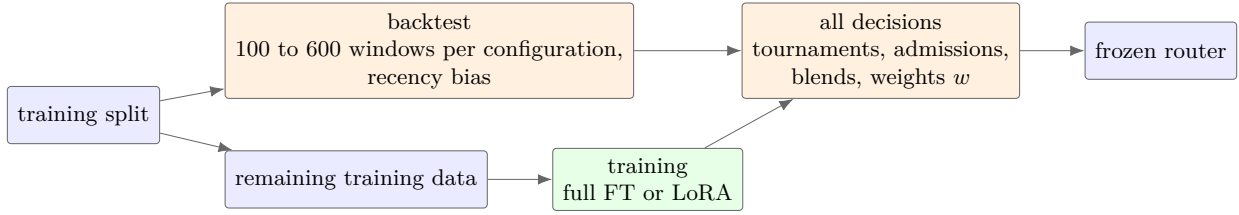

\paragraph{A backtest carved from the training split.} All decisions are
taken on context-plus-horizon windows sliced from the end of the training
split. Recent windows resemble the conditions the forecaster will face,
and on long series old regimes mislead selection. Windows are cut at
several positions per series, hourly regimes are identified, and each
configuration receives roughly 100 to 600 backtest windows. Candidates
are scored there as the benchmark scores them, with MASE computed on the
real context before any padding, and with the CRPS approximated by the
mean pinball loss $\rho_q(y,\hat y_q)=\max(q(y-\hat y_q),(q-1)(y-\hat
y_q))$ over $q\in\{0.1,\dots,0.9\}$ \citep{koenker1978quantile}. The
training data not consumed by the backtest trains the specialists
(Figure~\ref{fig:protocol}).

\paragraph{Validating the selector itself.} Before trusting a selection
rule we evaluate it by a temporal meta-backtest. The rule selects on the
older backtest windows and its choice is scored on the recent ones, over
hundreds of draws. The split is temporal, following
\citet{bergmeir2012crossvalidation} and \citet{cerqueira2020evaluating}
one level up, from models to selection rules. All rule iteration in this
work, on margins, tie-breaks and gate thresholds, was carried out inside
this protocol, at no cost in evaluation data.

\section{Building the experts}
\label{sec:experts}

The design principle of this section is that the value of a fine-tune is
decided before training starts, by how its data is prepared. Each
preparation rule below is cheap and explicit, and every candidate it
produces faces the admission rule of Definition~\ref{def:admission}.

\paragraph{Cleaning.} Every corpus is filtered series by series before
training. Degenerate series with near-zero variance, constant or
few-valued, crush the loss while teaching nothing. Scale outliers from
failing sensors or mixed units can dominate the gradient on their own.
Corrupted stretches are truncated instead of imputed. Per-source volume
caps in fusions stop any one dataset from drowning the others. Cleaning is
contextual. On a 21-series weather dataset, fusing with a large
neighboring corpus drowned the signal and the cleaned solo won, and the
opposite held for data-poor datasets. The rule that emerged is that a
data-rich configuration wants a dedicated specialist on cleaned data,
while a data-poor configuration wants an enriched family.

\paragraph{Dosed fusion.} For data-poor configurations the target's
training data is mixed with that of sibling configurations of the same
domain or frequency, with the target over-weighted, typically three parts
target to one part siblings. The siblings regularize and the target
remains the signal. Families are built by similarity, first frequency and
horizon, then domain, and entirely poor families borrow from rich
siblings. Rich families receive full fine-tuning and poor ones LoRA.
Unstable targets get several seeds, and a short run of 500 steps is
extended to 2{,}000 only if it already wins.

\paragraph{Controlled windows.} Training reproduces the evaluation
conditions. Windows are sampled with variable-length contexts, the target
is the full horizon or the next patch depending on the architecture, and
residuals are normalized robustly,
\[
z=\frac{x-\mathrm{med}(c)}{s(c)},\qquad
s(c)=\max\bigl(\mathrm{IQR}(c),\,0.05\cdot\mathrm{range}(c)\bigr),\qquad
|z|\le 20,
\]
without which the loss is dominated by outliers. The loss is the pinball
loss on the benchmark's own nine deciles, with sorted outputs, so the
model learns the distribution it will be judged on. Learning rates are
$10^{-4}$ for LoRA and $10^{-5}$ for full fine-tunes.

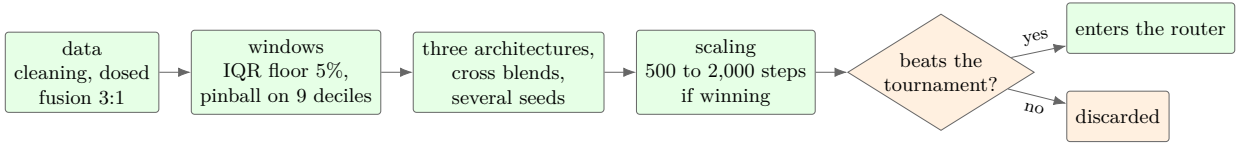
\begin{figure}[t]
\centering
\resizebox{\linewidth}{!}{%
\begin{tikzpicture}[node distance=3mm and 5mm]
\node[ex] (a) {data\\cleaning, dosed\\fusion 3:1};
\node[ex, right=of a] (w) {windows\\IQR floor 5\%,\\pinball on 9 deciles};
\node[ex, right=of w] (m) {three architectures,\\cross blends,\\several seeds};
\node[ex, right=of m] (s) {scaling\\500 to 2{,}000 steps\\if winning};
\node[dec, right=of s, aspect=1.6] (v) {beats the\\tournament?};
\node[ex, right=of v, yshift=7mm] (y) {enters the router};
\node[tn, right=of v, yshift=-7mm] (n) {discarded};
\draw[arr] (a) -- (w); \draw[arr] (w) -- (m); \draw[arr] (m) -- (s); \draw[arr] (s) -- (v);
\draw[arr] (v) -- node[above, sloped, font=\scriptsize] {yes} (y);
\draw[arr] (v) -- node[below, sloped, font=\scriptsize] {no} (n);
\end{tikzpicture}}
\caption{The expert pipeline. Cleaning and dosed fusion, controlled windows
with robust normalization and pinball loss, three architectures with cross
blends and conditional scaling, then the admission rule.}
\label{fig:pipeline}
\end{figure}

\paragraph{Three architectures and cross blends.} Chronos-2, TiRex and
Toto 2.5B are specialized, each with its native recipe.
Blends are sorted quantile means of their members, and \nBlendMixed{} of
the \nBlendAll{} blends in the released table mix architectures. Each candidate is a checkpoint of a few megabytes trained
in minutes, and by Definition~\ref{def:admission} it enters the router
only by beating the tournament on the backtest
(Figure~\ref{fig:pipeline}).

\section{The tournament and its guard rails}
\label{sec:tournament}

\paragraph{Stage 1, a selection defended against its own biases.} On
configurations without a designated specialist, every pool member is
scored on the backtest windows and the winner becomes the served default.
Three rules make this selection robust. A dual criterion requires both
MASE and CRPS, because members with degenerate quantiles, nine identical
deciles, can win MASE alone while being probabilistically useless. An
asymmetric anti-memorization margin handles locally trained candidates,
whose backtest scores are flattered because they saw these series during
training. Such a candidate is admitted only if $S(\mathrm{local}) <
(1-\delta)\cdot\min_{\mathrm{generic}} S$ with $\delta=12\%$, and only if
it is not worse on CRPS. Near-ties, with a gap below 3\%, are settled on
CRPS. When the blend of the pool's two largest architectures is within 5\% of
the winner, the blend is served instead, because a win by a hair is usually
backtest noise and blending is the safer choice under uncertainty.

\paragraph{Stage 2, a conservative per-window gate.} At prediction time
each window is checked by a lightweight backtest of its own context on two
segments, the final segment of horizon length and the middle segment. A
window is rerouted only if all conditions agree. The default's error must
exceed a threshold, $\tau=1.5$ for foundation models and $\tau=3$ for
naive references. The challenger must be better by at least 10\% on
average, better on each segment, and better on CRPS. A regime-transition
guard blocks any switch when the recent dormancy of the context differs
from what the segments saw, as at dawn and dusk of hourly series, where
the segment signal inverts. The gate is designed to switch rarely.

\paragraph{Safety nets for designated experts.} Configurations served by
a specialist carry a distinct anti-catastrophe gate with high thresholds,
$\tau=5$ to $10$. It intervenes only on flagrant per-window failure, in
which case the window falls back to the generic pool, and it never trims a
healthy specialist. An alignment guard verifies by correlation that
imported predictions from caches or external wrappers are attached to the
right windows. Plausible-looking but misaligned outputs are the worst kind
of error, since aggregate metrics hide them, and they are rejected when
the median correlation falls below $0.30$. Over our campaigns this guard
fired three times, each time on a real misalignment, and never on an
aligned import.

\section{Evaluation protocol}
\label{sec:evalproto}

\paragraph{Harness.} All reported scores come from the official GIFT-Eval
harness with the leaderboard parameters, which is the forecast evaluation
routine of gluonts \citep{alexandrov2020gluonts}. We
validated our installation by reproducing a published reference model to
three decimals. Every submitted line is recomposed from a saved artifact,
weights, adapters or verified quantiles, and checked for equality on MASE
and weighted quantile loss before submission.

\paragraph{Leaderboard snapshot.} The public leaderboard publishes one
per-configuration result file per entry. On 2026-09-14 we downloaded every
result file and every entry description from the leaderboard repository.
Entries with a result file covering all 97 configurations number
\nPublic. For each configuration we rank all entries by MASE and by
weighted quantile loss, ties sharing the lowest rank, and we average the
ranks over the 97 configurations. Our submission is inserted into this set as one more entry and ranked by
the same rule. Every pool member declares no training overlap with the
test sets in its leaderboard description, and our own training touches
only the provided training splits. The leaderboard is living,
so every position in this paper is dated. The snapshot itself is part of
the release (Section~\ref{sec:repro}).

\section{Results}
\label{sec:results}

\begin{figure}[t]
\centering
\includegraphics[width=0.95\linewidth]{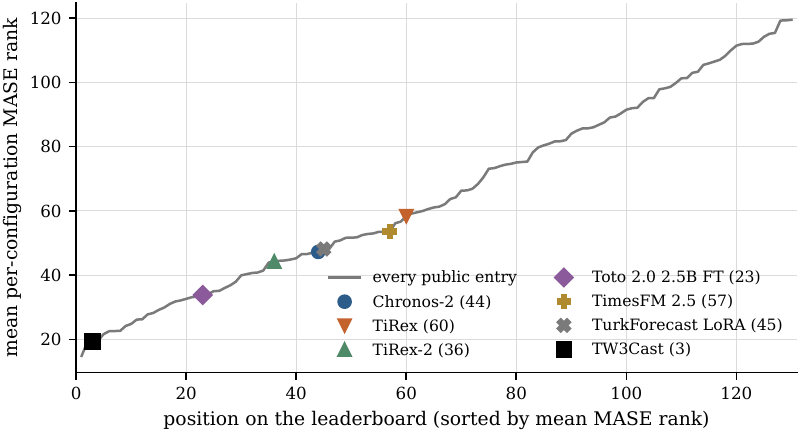}
\caption{Mean per-configuration MASE rank of the \nEntries{} entries of
Table~\ref{tab:leaderboard}, sorted by position, with the six members of
the tournament pool and TW3Cast marked; the number in parentheses is the
position. Script \code{make\_figures.py}.}
\label{fig:leaderboard}
\end{figure}

\paragraph{Result 1 (standing on the leaderboard).} Inserted among the
\nPublic{} public entries, TW3Cast reaches position \ourPos{} of
\nEntries{} by mean MASE rank, with a mean rank of \ourRank. Every entry above it is declared as agentic by its authors, the category
described in Section~\ref{sec:related}, so TW3Cast is the best entry that
runs no agent and no language model. By mean weighted
quantile loss rank the system is at position \ourWqlPos, with a mean rank
of \ourWqlRank, so its calibration lags its point accuracy.
Table~\ref{tab:leaderboard} gives the top of the leaderboard and
Figure~\ref{fig:leaderboard} the full distribution.

\begin{table}[t]
\centering
\footnotesize
\setlength{\tabcolsep}{4.5pt}
\begin{tabular}{rlrll}
\toprule
Position & Entry & Mean MASE rank & Declared type & Test leakage \\
\midrule
1 & STRIDE\_w\_Synapse & 14.9 & agentic & No \\
2 & EXAONE-Forecast-Agent & 19.2 & agentic & No \\
3 & TW3Cast (this work) & 19.4 & router & No \\
4 & LS-MoE & 19.7 & agentic & No \\
5 & LS-Agent & 21.7 & agentic & No \\
6 & Falcon-Agent & 22.5 & agentic & No \\
7 & CastStar & 22.6 & agentic & No \\
8 & limix\_moe & 22.7 & agentic & No
 \\
\bottomrule
\end{tabular}
\caption{Top of the GIFT-Eval leaderboard by mean per-configuration MASE
rank on 2026-09-14, computed by \code{make\_figures.py} from the
per-configuration result files of the \nPublic{} public entries with
complete results, with the submitted TW3Cast result file inserted as one
more entry and ranked by the rule of Section~\ref{sec:evalproto}.
Declared type and leakage are copied from each entry's own description
file on the leaderboard. The snapshot of the public files, our result
file and the script are in the release.}
\label{tab:leaderboard}
\end{table}

\begin{figure}[t]
\centering
\includegraphics[width=0.95\linewidth]{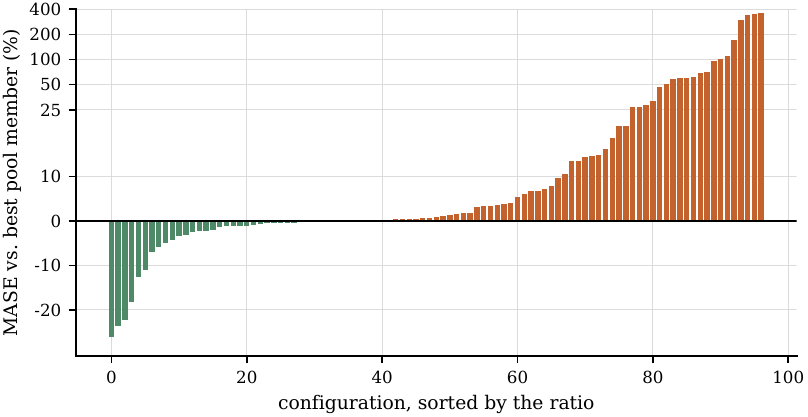}
\caption{Per-configuration MASE of TW3Cast relative to the best pool
member on that configuration, in percent, over the 97 configurations
sorted by the ratio. Negative bars are configurations where the system
beats the per-configuration oracle over the pool. Script
\code{make\_figures.py}.}
\label{fig:perconfig}
\end{figure}

\paragraph{Result 2 (components alone do not explain the position).} The
building blocks are plain public foundation models, each of which is also
a public leaderboard entry, and Table~\ref{tab:components} gives their
own standing. The best of them, \bestPoolName, sits at position
\bestPoolPos{} with a mean MASE rank of \bestPoolRank, and the weakest
pool member has a mean rank of \worstPoolRank. A harder comparison is the
per-configuration oracle that picks, on each configuration, the pool
member with the lowest test MASE, a choice that requires test knowledge.
Against that oracle the system is better on \nBeatBest{} of the 97
configurations, within five percent on \nWithinFive, and more than ten
percent worse on \nWorseTen{} (Figure~\ref{fig:perconfig}). The gap
between the components and the complete system is created by data
preparation, training and guarded selection, since nothing else enters
the system.

\begin{table}[t]
\centering
\footnotesize
\begin{tabular}{lrrrl}
\toprule
Pool member & Position & MASE rank & WQL rank & Declared type \\
\midrule
Chronos-2 & 44 & 47.2 & 47.6 & pretrained \\
TiRex & 60 & 58.4 & 51.1 & zero-shot \\
TiRex-2 & 36 & 44.2 & 42.7 & pretrained \\
Toto 2.0 2.5B FT & 23 & 33.8 & 33.1 & fine-tuned \\
TimesFM 2.5 & 57 & 53.5 & 53.1 & zero-shot \\
TurkForecast LoRA & 45 & 48.1 & 47.2 & fine-tuned
 \\
\bottomrule
\end{tabular}
\caption{Standing of each tournament pool member as its own public
leaderboard entry, in the same field and with the same rank computation
as Table~\ref{tab:leaderboard}. The WQL column is the same average on
weighted quantile loss. TiRex-2 is listed under its pretrained entry, the
one used in the pool.}
\label{tab:components}
\end{table}

\paragraph{Result 3 (the tournament and the specialists each earn their
place).} Table~\ref{tab:ablation} ranks, in the same public field, each
pool member served alone, a variant in which the tournament serves every
configuration with no designated specialist, and the released router.
Served on every configuration the tournament reaches a mean MASE rank of
\tourOnlyRank, behind the best single member at \bestPoolRank. On the \nTournamentConfigs{} configurations that
the released router actually assigns to the tournament, the same
tournament reaches \tourOnRankSub{} where the best member reaches
\bestPoolRankSub. On the other \nOther{} configurations the designated
specialist or blend beats the tournament-only serving on \nSpecBeatsTour,
with a median MASE gain of \medGainSpec{} percent. The router is worth
more than either of its halves because it assigns each half the
configurations it wins.

\begin{table}[t]
\centering
\footnotesize
\begin{tabular}{lrr}
\toprule
Variant & All 97 & Tournament configurations (\nTournamentConfigs) \\
\midrule
Chronos-2 & 47.2 & 42.6 \\
TiRex & 58.4 & 63.2 \\
TiRex-2 & 44.2 & 41.5 \\
Toto 2.0 2.5B FT & 33.8 & 28.1 \\
TimesFM 2.5 & 53.5 & 58.8 \\
TurkForecast LoRA & 48.1 & 51.4 \\
Tournament on every configuration & 38.0 & 19.8 \\
Full system (released router) & 19.4 & 19.7
 \\
\bottomrule
\end{tabular}
\caption{Mean per-configuration MASE rank of each variant in the field of
Table~\ref{tab:leaderboard}, over all 97 configurations and over the
\nTournamentConfigs{} configurations that the released router serves by
tournament. Pool members are their public entries and keep the ranks of
Table~\ref{tab:components}. The tournament-only variant is scored from
its released per-configuration grid, inserted into the same field as one
more entry; on the tournament configurations that grid matches the submission to
within 0.2 percent. Script \code{make\_figures.py}.}
\label{tab:ablation}
\end{table}

\paragraph{Result 4 (where training won and where it lost).} The
admission rule makes negative results cheap and visible. On several
configurations, among them two short-horizon business-monitoring datasets
and an hourly weather dataset, every fine-tuned candidate we trained
across the three architectures lost to the tournament on the backtest and
was discarded. On those configurations the tournament's cross-architecture
blend with the per-window gate is the served mode. Each discarded
candidate cost minutes of GPU time and changed nothing in the released
system. The same mechanism captured the wins. On a 10-minute weather
configuration a cleaned-solo LoRA of Chronos-2 beat the tournament and was
designated, and on the \nWithSpecialist{} configurations served by or
with a specialist the designated expert beat the tournament on the
backtest by construction.

\section{Deployability and accretion}
\label{sec:deploy}

The shipped system is two small files and a set of checkpoints. Inference
is a table lookup, the experts' forward passes and a sorted weighted
average of quantiles, with no agent and no language model in the loop.

The router improves by accretion. A candidate expert is a LoRA adapter of
a few megabytes, trained in 500 to 2{,}000 steps, two to seven minutes on
a single GPU. Admission requires beating the tournament on the train-side
backtest, so adding an expert never degrades the system on its selection
criterion, and a failed candidate costs minutes and changes nothing. No global retraining ever occurs. Integrating, replacing
or retiring an expert is a one-row change in the table.

The same property drives adaptation beyond the benchmark, because nothing
in the method depends on configuration names. On a new data source the
tournament builds its backtest on the available history and serves a
default with the same guard rails, with no training at all. When
specialization is worth trying, the pipeline of Section~\ref{sec:experts}
applies to the source's history and the admission rule arbitrates. The
router grows wherever cheap training wins and falls back to its tournament
wherever it does not.

\section{Threats to validity}
\label{sec:threats}

\paragraph{A living leaderboard.} Positions move with every new
submission. Every rank in this paper is therefore computed from a dated
snapshot of the public per-configuration files, which the release
contains, and the rank computation is a short script rather than a
reading of the leaderboard page. The snapshot fixes the comparison set but
does not fix the future, and a later reader should expect a different
position.

\paragraph{Recency bias as an assumption.} The backtest samples the end
of the training split on purpose. This is an assumption about regime
continuity, not a theorem. The temporal meta-backtest is built on the same assumption, and a source
with an abrupt regime change at the train and test boundary would defeat
both.

\paragraph{Selection flexibility.} Any system with margins and thresholds
can overfit its own validation data. All such constants
(Appendix~\ref{app:constants}) were chosen inside the temporal
meta-backtest and then frozen, and none was tuned on official evaluation
results. The protection is procedural, and it does not remove the
dependence of the constants on this benchmark's mix of frequencies and
domains.

\paragraph{Calibration.} The system is selected and admitted under a
dual MASE and CRPS criterion, yet its weighted quantile loss rank
(position \ourWqlPos) lags its MASE rank (position \ourPos). The dual criterion eliminates poorly calibrated winners but does not
select for calibration. A CRPS-first admission rule is the obvious
variant and is not evaluated here.

\paragraph{Serialization drift.} One serialization path of a base library
drifted by under one percent between the in-memory model and its reloaded
checkpoint. Only reloaded artifacts may produce published numbers under
our rule, one designated expert was re-scored from its reloaded checkpoint
for this reason, and checkpoint formats with bit-exact reload are
preferred throughout. One expert whose seed-averaged weights were not
retained is shipped as its verified forecast quantiles instead, which
reproduces its forecasts but not its training.

\section{Reproducibility}
\label{sec:repro}

The code repository\footnote{\url{https://github.com/TW3-Partners-OS/TW3-Cast}}
contains the routing table \code{router.csv}, the expert index
\code{experts.json}, the pinned revisions of the six base models
\code{base\_models.json}, and a reference inference script
\code{predict.py}. Expert checkpoints are distributed with the repository under their expert
identifiers, and the one expert shipped as quantiles is loaded from its
quantile file.
The paper's own folder in the repository contains the dated snapshot
of the public per-configuration result files, the submitted TW3Cast
result file, the per-configuration grid of the tournament-only variant,
and the script \code{make\_figures.py} that regenerates every table,
figure and number of this paper from them.

Two regimes apply. The exact regime regenerates every table, figure and
number of this paper from the release alone, with no model call, in
seconds. The fresh regime regenerates our result file by running the inference
script on each configuration and scoring it with the official harness,
which requires the base models at their pinned revisions, the expert
checkpoints and one GPU. The tournament procedure for the \nTournament{} configurations it serves
is described in Section~\ref{sec:tournament}.

\section{Conclusion}
\label{sec:conclusion}

A frozen table of mixture weights over lightly fine-tuned experts, fed by
an explicit data-preparation methodology and defended by guarded
train-side selection, reaches position \ourPos{} of \nEntries{} on
GIFT-Eval by mean MASE rank on 2026-09-14 once inserted among the
\nPublic{} public entries, with no agent and no language model at inference.
The design optimizes for properties that large monolithic forecasters
lack. Every routing row traces to a measured decision, admission by
beating the tournament makes each upgrade individually verified at minutes
per candidate, and the same machinery produces a router for any new set of
series. The accretion property is the consequence we expect to matter
most in practice, because it turns the maintenance of a forecasting system
from periodic retraining into a stream of small, cheap, verified changes.

\bibliographystyle{plainnat}
\bibliography{refs}

\appendix

\section{Constants and thresholds}
\label{app:constants}

\begin{table}[ht]
\centering
\footnotesize
\begin{tabular}{llll}
\toprule
Mechanism & Constant & Value & Set by \\
\midrule
Anti-memorization margin & $\delta$ & 12\% & meta-backtest \\
Near-tie CRPS tie-break & gap & $<3\%$ & meta-backtest \\
Blend fallback & gap to winner & $<5\%$ & meta-backtest \\
Per-window gate & $\tau$ (foundation / naive) & 1.5 / 3.0 & meta-backtest \\
Per-window gate & challenger advantage & $\ge 10\%$ & meta-backtest \\
Anti-catastrophe gate & $\tau$ & 5 to 10 & meta-backtest \\
Alignment guard & median correlation & $\ge 0.30$ & incident analysis \\
Normalization & scale floor & $0.05\cdot\mathrm{range}$ & training stability \\
Normalization & residual clip & $|z|\le 20$ & training stability \\
Fine-tuning & LoRA / full learning rate & $10^{-4}$ / $10^{-5}$ & standard \\
Fine-tuning & steps (short / extended) & 500 / 2{,}000 & conditional scaling \\
Backtest & windows per configuration & 100 to 600 & budget \\
\bottomrule
\end{tabular}
\caption{All fixed constants of the released system. Every constant was
chosen inside the train-side protocols of Sections~\ref{sec:protocol}
and~\ref{sec:tournament} and then frozen.}
\label{tab:constants}
\end{table}

\section{The released routing table}
\label{app:router}

Tables~\ref{tab:router} and~\ref{tab:routerb} list the served mode of
every configuration as read from \code{router.csv}. Abbreviations are C2 for Chronos-2, TFM for
TimesFM 2.5, Toto for Toto 2.0 2.5B FT, Toto-uni for a univariate Toto member, Turk for the TurkForecast LoRA of Chronos-2, and TiRex and
TiRex-2 for the two TiRex releases.

\begin{table}[p]
\centering
\footnotesize
\setlength{\tabcolsep}{4pt}
\begin{tabular}{ll}
\toprule
Configuration & Served by \\
\midrule
bitbrains\_fast\_storage/5T/long & tournament \\
bitbrains\_fast\_storage/5T/medium & E01\,(Toto) \\
bitbrains\_fast\_storage/5T/short & tournament \\
bitbrains\_fast\_storage/H/short & E02\,(Toto) \\
bitbrains\_rnd/5T/long & Toto + C2 + Turk \\
bitbrains\_rnd/5T/medium & Toto + C2 + Turk \\
bitbrains\_rnd/5T/short & tournament \\
bitbrains\_rnd/H/short & C2 + Turk \\
bizitobs\_application/10S/long & tournament \\
bizitobs\_application/10S/medium & tournament \\
bizitobs\_application/10S/short & tournament \\
bizitobs\_l2c/5T/long & E03\,(C2) \\
bizitobs\_l2c/5T/medium & E03\,(C2) \\
bizitobs\_l2c/5T/short & tournament \\
bizitobs\_l2c/H/long & E42\,(Toto) \\
bizitobs\_l2c/H/medium & E42\,(Toto) \\
bizitobs\_l2c/H/short & Toto + C2 + TFM \\
bizitobs\_service/10S/long & tournament \\
bizitobs\_service/10S/medium & Toto + C2 + TFM \\
bizitobs\_service/10S/short & tournament \\
car\_parts/M/short & C2 + TFM + TiRex-2 \\
covid\_deaths/D/short & tournament \\
electricity/15T/long & E09\,(Toto) + C2 + TiRex \\
electricity/15T/medium & E09\,(Toto) + C2 + Turk \\
electricity/15T/short & tournament \\
electricity/D/short & E04\,(C2) + Toto \\
electricity/H/long & E05\,(C2) \\
electricity/H/medium & E06\,(Toto) + C2 + E07\,(C2) \\
electricity/H/short & Toto + C2 + TiRex \\
electricity/W/short & E08\,(C2) \\
ett1/15T/long & Turk + C2 + E09\,(Toto) + TiRex \\
ett1/15T/medium & E10\,(C2) + C2 \\
ett1/15T/short & Toto + C2 + TiRex \\
ett1/D/short & E11\,(C2) + C2 \\
ett1/H/long & TiRex + E12\,(TiRex) + TiRex-2 \\
ett1/H/medium & E13\,(C2) + Toto-uni \\
ett1/H/short & tournament \\
ett1/W/short & TFM + TiRex-2 + C2 \\
ett2/15T/long & E14\,(C2) + Turk \\
ett2/15T/medium & E10\,(C2) + E13\,(C2) \\
ett2/15T/short & Toto + C2 + TFM \\
ett2/D/short & E04\,(C2) + E15\,(C2) \\
ett2/H/long & Turk + TFM + E16\,(Toto) \\
ett2/H/medium & E13\,(C2) + Toto-uni \\
ett2/H/short & E17\,(TiRex) + C2 \\
ett2/W/short & E18\,(Toto) \\
hierarchical\_sales/D/short & Toto + C2 + TiRex \\
hierarchical\_sales/W/short & E19\,(C2) \\
hospital/M/short & tournament
 \\
\bottomrule
\end{tabular}
\caption{The released routing table, first half of the 97 configurations
of \code{router.csv} in alphabetical order, with the members of each
blend. Members named E\emph{nn} are designated specialists, with their
architecture family in parentheses; other names are base models of the
tournament pool at their pinned revisions; blend members carry equal
weights. Generated by \code{make\_figures.py}.}
\label{tab:router}
\end{table}

\begin{table}[p]
\centering
\footnotesize
\setlength{\tabcolsep}{4pt}
\begin{tabular}{ll}
\toprule
Configuration & Served by \\
\midrule
jena\_weather/10T/long & tournament \\
jena\_weather/10T/medium & E44\,(C2) \\
jena\_weather/10T/short & tournament \\
jena\_weather/D/short & E20\,(TiRex) + TFM \\
jena\_weather/H/long & tournament \\
jena\_weather/H/medium & E43\,(C2) + TFM + C2 \\
jena\_weather/H/short & tournament \\
kdd\_cup\_2018/D/short & Toto + TFM + Turk \\
kdd\_cup\_2018/H/long & E22\,(TiRex) + TiRex \\
kdd\_cup\_2018/H/medium & E22\,(TiRex) + TiRex \\
kdd\_cup\_2018/H/short & tournament \\
loop\_seattle/5T/long & tournament \\
loop\_seattle/5T/medium & tournament \\
loop\_seattle/5T/short & Toto + C2 + TFM \\
loop\_seattle/D/short & E23\,(TiRex) + TFM \\
loop\_seattle/H/long & Toto + C2 + TFM \\
loop\_seattle/H/medium & E24\,(Toto) \\
loop\_seattle/H/short & E27\,(Toto) + TFM + C2 \\
m4\_daily/D/short & E25\,(TiRex) + E26\,(TiRex) \\
m4\_hourly/H/short & E27\,(Toto) + TiRex + TFM \\
m4\_monthly/M/short & tournament \\
m4\_quarterly/Q/short & E32\,(C2) + Toto \\
m4\_weekly/W/short & Toto + TFM \\
m4\_yearly/A/short & tournament \\
m\_dense/D/short & Turk + E28\,(Toto) + TFM \\
m\_dense/H/long & Toto + C2 + TFM \\
m\_dense/H/medium & Toto + C2 + TFM \\
m\_dense/H/short & E29\,(TiRex) + C2 + E27\,(Toto) \\
restaurant/D/short & E30\,(C2) + Turk + E31\,(TiRex) \\
saugeen/D/short & C2 + TFM \\
saugeen/M/short & Toto + C2 + TFM \\
saugeen/W/short & Toto + Turk \\
solar/10T/long & Toto + C2 + Turk \\
solar/10T/medium & tournament \\
solar/10T/short & Toto + C2 + Turk \\
solar/D/short & tournament \\
solar/H/long & TiRex + E33\,(TiRex) + E34\,(TiRex) \\
solar/H/medium & E35\,(TiRex) \\
solar/H/short & tournament \\
solar/W/short & E19\,(C2) \\
sz\_taxi/15T/long & E36\,(TiRex) + Toto \\
sz\_taxi/15T/medium & E36\,(TiRex) + Toto \\
sz\_taxi/15T/short & tournament \\
sz\_taxi/H/short & E27\,(Toto) + E37\,(C2) \\
temperature\_rain/D/short & E41\,(C2) \\
us\_births/D/short & tournament \\
us\_births/M/short & E38\,(C2) + E39\,(C2) \\
us\_births/W/short & E40\,(Toto)
 \\
\bottomrule
\end{tabular}
\caption{The released routing table, second half, same conventions as
Table~\ref{tab:router}.}
\label{tab:routerb}
\end{table}

\end{document}